%% file: paper.tex
\documentclass{article}
\usepackage{spconf}
\usepackage[utf8]{inputenc} 
\usepackage[T1]{fontenc}    
\usepackage{hyperref}       
\usepackage{url}            
\usepackage{booktabs}       
\usepackage{amsfonts}       
\usepackage{nicefrac}       
\usepackage{microtype}      
\usepackage[algoruled,boxed,lined]{algorithm2e}
\usepackage{graphicx}
\usepackage{pifont}
\usepackage[tight,footnotesize]{subfigure}
\usepackage{amsmath} 
\usepackage{amssymb}  
\usepackage{xcolor}
\usepackage{colortbl}
\usepackage{array,colortbl,ctable}
\usepackage{bbm}
\usepackage{subfigure}
\newcolumntype{P}[1]{>{\centering\arraybackslash}m{#1}}

\title{UNDERSTANDING DEEP NEURAL NETWORKS THROUGH INPUT UNCERTAINTIES}

\name{Jayaraman J. Thiagarajan$^{\ddagger}$\thanks{This work was performed under the auspices of the U.S. Department of Energy by Lawrence Livermore National Laboratory under Contract DE-AC52-07NA27344.}, Irene Kim$^{\dagger \ddagger}$, Rushil Anirudh$^\ddagger$ and Peer-Timo Bremer$^\ddagger$}
\address{$^\dagger$ University of California Davis, $^\ddagger$Lawrence Livermore National Laboratory \\
Email:\{jjayaram@llnl.gov, imkkim@ucdavis.edu, anirudh1@llnl.gov, bremer5@llnl.gov\}}

\begin{document}
	%
	\maketitle
	\begin{abstract}
          Techniques for understanding the functioning of complex
          machine learning models are becoming increasingly popular,
          not only to improve the validation process, but also to extract
          new insights about the data via exploratory analysis. Though
          a large class of such tools currently exists, most assume
          that predictions are point estimates and use a 
          sensitivity analysis of these estimates to
          interpret the model. Using lightweight probabilistic
          networks we show how including prediction uncertainties in
          the sensitivity analysis leads to: (i) more robust and
          generalizable models; and (ii) a new approach for model
          interpretation through uncertainty decomposition. In
          particular, we introduce a new regularization that takes
          both the mean and variance of a prediction into account and
          demonstrate that the resulting networks provide improved generalization to unseen data. Furthermore, we propose a
          new technique to explain prediction uncertainties through 
          uncertainties in the input domain, thus providing new ways to
          validate and interpret deep learning models. 
		
	\end{abstract}
	
	\section{Introduction}
	\input{intro}

	\section{Lightweight Probabilistic Networks}
	\label{sec:lw}
	\input{lightweight}
	\section{Proposed Approach}
	\input{approach}

	\section{Experiments}
	\input{exps}
	
	\nocite{gast2018lightweight}
	\bibliographystyle{IEEEbib}
	\bibliography{refs}
	
\end{document}

%% file: intro.tex
Machine learning techniques, such as deep neural networks (DNNs), have become a central component of analytics pipelines in science and engineering. With this widespread adoption, it is critical to verify that the superior prediction performance arises from meaningful patterns rather than from artifacts or biases in the data. Consequently, techniques that can enable understanding of what a model has learned are an integral part of validation processes~\cite{zhou2017deep,ackerman2017drive}. While the notion of \textit{interpretability} has several definitions throughout the literature, we restrict our focus on understanding model predictions in terms of simple constructs that are easily actionable, most notably the input features. For example, one would like to answer questions, such as: \textit{Which input features helped the decision? How confident is the model about a decision?} etc.  

\noindent In conventional statistical modeling these questions are typically answered through uncertainty quantification of a pre-trained model. However, adopting a fully Bayesian inferencing pipeline for DNNs, i.e.\ modeling every neuron as a statistical distribution, is not feasible in practice~\cite{graves2011practical}. Instead, most DNN architectures only provide point estimates, and thus may appear highly confident of their predictions even while making mistakes. For example, in a classification task for an out-of-distribution test sample, a trained DNN can still erroneously produce a \textit{softmax} distribution concentrated around one of the classes. More importantly, computing sensitivites from point estimates~\cite{montavon2017methods} to expain a given decision scores are entirely based on local gradients $\left(\partial{f}/ \partial x_j\right)^2$. While this does identify the feature(s) $j$ of a sample $\mathbf{x}$ that will lead to maximal changes in the prediction $f(\mathbf{x}; \boldsymbol{\theta})$ it says little about how these features affect the prediction uncertainty. Depending on the application, a feature with larger sensitivity but comparatively small effect on prediction uncertainty may be less concerning than a low sensitivity feature that leads to large change in uncertainties. This additional level of detail in sensitivity analysis opens up a wide range of possibilities in feature understanding and selection, which has not been possible until now. 

\noindent Here, we build upon recent efforts to develop tractable techniques to approximate prediction uncertainties in DNNs~\cite{bendale2016towards, gal2016dropout, guo2017calibration, gast2018lightweight}. Note, there are two forms of predictive uncertainties in DNNs: \textit{epistemic} uncertainty, also known as model uncertainty that can be explained away given enough training data, and \textit{aleatoric} uncertainty, which depends on noise or randomness in the input sample. We adopt the latter approach, similar to~\cite{gast2018lightweight}, that produces both mean and variance estimates for the prediction, assuming some prior distribution on the inputs. We show that including both mean and variance in the sensitivity analysis produces more robust explanations, and when used as regularizers, these lead to better generalization. Finally, we introduce a new technique to determine which feature the model suspects to contribute maximally to the prediction uncertainty. This not only provides a novel approach to validate DNNs but also a new technique to interpret model decisions. 
For example, finding that a model assigns most of the uncertainties to otherwise reliable outputs suggest problems in either the training process or the input data. Using benchmark regression datasets, we demonstrate the effectiveness of the proposed approaches to build robust, yet interpretable, predictive models.

%% file: lightweight.tex
Before we describe the proposed approach, we briefly review the formulation of lightweight networks~\cite{gast2018lightweight}. For a given input matrix $\mathbf{X} = [\mathbf{x}_1, \mathbf{x}_1, \cdots, \mathbf{x}_N ]^T$ and their corresponding outputs, $\mathbf{y} = [y_1,y_2,\cdots y_N]$, our goal is to infer a predictive model $f: \mathbf{x} \rightarrow y$, with parameters $\boldsymbol{\theta}$. LPN performs propagation of aleatoric uncertainty through the network, wherein each input sample is modeled using an independent univariate Gaussian distribution in each of the dimensions. The propagation of uncertainties is carried out using \textit{Assumed Density Filtering} \cite{boyen1998tractable}, where each layer is implemented as filtering of the input distribution to obtain a transformed Gaussian distribution with diagonal covariance. In contrast to other Bayesian deep learning formulations, where the model parameters are assumed to be stochastic, LPN keeps the model parameters deterministic. 


For a sample $\mathbf{x}$, the joint density of all activations is 
$$p(\mathbf{z}^{(0:\ell)}) = p(\mathbf{z}^{(0)}) \prod_{l = 1}^{\ell}p(\mathbf{z}^{(l)}|\mathbf{z}^{(l-1)}),$$where $\ell$ denotes the number of layers and $\mathbf{z}$'s are the activations. With the independent Gaussian assumption, for an $l^{\text{th}}$ layer,
$$p(\mathbf{z}^{(l)}) = \prod_j \mathcal{N}(\mu_j^{(l)}, \nu_j^{(l)}),$$where $j$ is the index of a neural unit. For simplicity, we denote this as $p(\mathbf{z}^{(l)}) = \mathcal{N}(\boldsymbol{\mu}^{(l)}, \boldsymbol{\nu}^{(l)})$. At the input layer, we set $\boldsymbol{\mu}^{(0)} = \mathbf{x}$ and $\boldsymbol{\nu}^{(0)} = \boldsymbol{\sigma}$, where $\boldsymbol{\sigma}$ is a prior on the aleatoric uncertainties. For efficient implementation, we can obtain analytical expressions for the filtering operation corresponding to commonly employed layers in neural network architectures.


\noindent \textbf{Dense layer:} For a fully connected layer with weights $\mathbf{W}$ and bias $\mathbf{b}$, the input distribution $\mathcal{N}(\boldsymbol{\mu}, \boldsymbol{\nu})$ can be filtered to produce a Gaussian with mean and variance
\begin{align}
\nonumber \boldsymbol{\mu}_{fc} = \mathbf{W}\boldsymbol{\mu};\text{    } \boldsymbol{\nu}_{fc} = (\mathbf{W} \circ \mathbf{W}) \boldsymbol{\nu}.
\end{align}Here, $\circ$ denotes element-wise product.

\noindent \textbf{ReLU Activation:} The filtering corresponding to the ReLU activation produces
\begin{align}
\nonumber & \boldsymbol{\mu}_{relu}(\boldsymbol{\mu}, \boldsymbol{\nu}) = \boldsymbol{\mu} \Phi\left(\frac{\boldsymbol{\mu}}{\sqrt{\boldsymbol{\nu}}}\right) + \sqrt{\boldsymbol{\nu}} \phi\left(\frac{\boldsymbol{\mu}}{\sqrt{\boldsymbol{\nu}}}\right),\\
\nonumber & \boldsymbol{\nu}_{relu}(\boldsymbol{\mu}, \boldsymbol{\nu}) = (\boldsymbol{\mu}^2 + \boldsymbol{\nu})\Phi\left(\frac{\boldsymbol{\mu}}{\sqrt{\boldsymbol{\nu}}}\right) + \boldsymbol{\mu}\sqrt{\boldsymbol{\nu}} \phi\left(\frac{\boldsymbol{\mu}}{\sqrt{\boldsymbol{\nu}}}\right) - \boldsymbol{\mu}_{relu}^2.
\end{align}Here, $\Phi$ and $\phi$ are standard normal and cumulative normal distributions respectively. These expressions show that in non-linear layers, mean and variance interact with each other.

\noindent \textbf{Leaky ReLU Activation:} Using the filtering expression for ReLU, we can derive mean and variance for leaky ReLU as
\begin{align}
\nonumber \boldsymbol{\mu}_{leaky\_relu}(\boldsymbol{\mu}, \boldsymbol{\nu}) &= \boldsymbol{\mu}_{relu}(\boldsymbol{\mu}, \boldsymbol{\nu}) - c  \boldsymbol{\mu}_{relu}(-\boldsymbol{\mu}, \boldsymbol{\nu}),\\
\nonumber \boldsymbol{\nu}_{leaky\_relu}(\boldsymbol{\mu}, \boldsymbol{\nu})  &=  \boldsymbol{\nu}_{relu}(\boldsymbol{\mu}, \boldsymbol{\nu}) + c^2  \boldsymbol{\nu}_{relu}(-\boldsymbol{\mu}, \boldsymbol{\nu}) \\
\nonumber  & + 2c \boldsymbol{\mu}_{relu}(\boldsymbol{\mu}, \boldsymbol{\nu}) \boldsymbol{\mu}_{relu}(-\boldsymbol{\mu}, \boldsymbol{\nu}).
\end{align}

\noindent \textbf{Dropout:} This is carried out by dropping each univariate normal distribution in a layer independently with a dropout rate $0<p<1$. Let $\mathbbm{1}_{\boldsymbol{\mu},p} = (b_1, b_2, ...., b_k)$ where $b_i$ are independent Bernoulli random variables with success probability $1-p$, and use $\boldsymbol{\mu} \circ \mathbbm{1}_{\boldsymbol{\mu},p}$ in lieu of $\boldsymbol{\mu}$, to perform dropout on the means. Subsequently, when a ReLU or leaky ReLU activation is applied, the filtering produces $0$ for both mean and variances, implying that the chosen neuron is dropped. 

%% file: approach.tex
In this section, we describe the proposed approach, that first estimates feature sensitivities using LPNs and refines model parameters using a novel loss function based on sensitivities. Next, we propose to study the input uncertainties in LPNs, with respect to degradation in the prediction uncertainty to gain a functional understanding of black-box models.

\subsection{Sensitivity Analysis With Probabilistic Networks}

\noindent \textbf{Architecture:} In this paper, we are interested in predicting a continuous response variable (i.e. regression), $\mathbf{y} \in \mathbb{R}^{N}$, using high-dimensional input features, $\mathbf{X} \in \mathbb{R}^{N \times d}$, where $d$ denotes the total number of input dimensions. Consequently, we follow the approach described in the previous section and construct a network based on assumed density filtering, comprising stacked dense layers, leaky ReLU activations and dropout (optional). Following  notations in Section \ref{sec:lw}, an input sample can be described by a set of independent Gaussians as follows:
$$p(\mathbf{x}_i) = \mathcal{N}(\mathbf{x}_i, \boldsymbol{\sigma_i}) =  \prod_{j=1}^d \mathcal{N}({x}_i^{(j)}, \sigma_i^{(j)}).$$Similarly, the prediction $\hat{y}_i = f(\mathbf{x}_i; \boldsymbol{\theta})$ from the model can be denoted as $\mathcal{N}(\hat{y}_i, \beta_i)$, where $\beta_i$ is the prediction variance. 

\noindent \textbf{Training:} With no prior knowledge about the input domain, we begin with a uniform uncertainty structure, $\sigma_i^{(j)} = \delta$, where $\delta > 0$ is a pre-defined constant (fixed at $\delta = 0.01$). For the actual training, we utilize the conditional likelihood based loss function, which is based on the general power exponential distribution family~\cite{gomez1998multivariate}. In particular, we minimize
\begin{equation}
\sum_i -\log p(y_i | \hat{y}_i, \beta_i) \propto \sum_i \log \beta_i + \left(\frac{(y_i - \hat{y}_i)^2}{\beta_i}\right)^k,
\label{eqn:loss}
\end{equation}where $k$ was fixed at $0.5$. In essence, the conditional log-likelihood amounts to the squared error weighted by its uncertainty along with a term that ensures that the prediction variance stays low. Upon training, the model $f$ can reduce this loss by either improving the mean prediction or by increasing the variance $\beta_i$ in the quest of improving $\hat{y}_i$. 


\noindent \textbf{Sensitivity Score:} Given the trained model, we measure feature sensitivities using an approach similar to~\cite{bach2015pixel}, but with the difference that we take into account both mean and variance estimates from the model. The Taylor decomposition method describes the model’s decision by decomposing the function value $f(\mathbf{x})$ as a sum of relevance scores, obtained using a first-order Taylor expansion of the function at some root point $\tilde{\mathbf{x}}$ such that $f(\tilde{\mathbf{x}}) = 0$. Extending the idea in~\cite{bach2015pixel}, we measure the relevance score for each input feature for a sample $\mathbf{x}_i$ as
\begin{equation}
\mathcal{R}^p_j(\mathbf{x}_i) = \left(x_i^{(j)}\frac{\partial \hat{y}_i}{\partial x_i^{(j)}}\right)^2 +  \left(x_i^{(j)} \frac{\partial \beta_i}{\partial x_i^{(j)}}\right)^2.
\label{eqn:sens}
\end{equation}A feature can be highly sensitive if its local variations can significantly alter the predicted mean or the variance.

\subsection{Explanation as Regularization}
In general, a neural network model $f$ can be considered to be \textit{explainable}, if one can identify a collection of interpretable features (e.g. a subset of input features) that maximally contribute to a particular decision. For example, the relevance scores in (\ref{eqn:sens}) can be used as a plausible explanation. We propose to utilize these explanations to actually regularize the network training process. In simpler terms, we aim to ensure that the model makes decisions for the right reasons (indicated by the explanations), in addition to producing the right answers~\cite{ross2017right}. More specifically, we refine the model parameters using a novel penalty term based on the estimated sensitivities -- we enforce the sensitivities for each sample to be more concentrated around the critical parameters, through conditional entropy. Hence, we refine the model parameters using the following objective:
\begin{equation}
\sum_i \log \beta_i + \left(\frac{(y_i - \hat{y}_i)^2}{\beta_i}\right)^k - \lambda h\left((\mathbf{x}_i) \ln  h(\mathbf{x}_i)\right),
\label{eqn:newloss}
\end{equation}where $h(\mathbf{x}_i)$ is a vector of relevance scores $\mathcal{R}^p_j(\mathbf{x}_i), \forall j$. The hyperparameter $\lambda$ was set to $1e-3$ in all experiments. From our experiments, we find that this refinement leads to much improved generalization, when compared to standard deterministic neural networks with similar configurations.

\subsection{Analysis of Input Uncertainties}
Basically, the input uncertainties can be used to convey the confidence on the input features. For example, these uncertainties can be related to the sampling distribution of the training data, i.e. heavily sampled regions can have higher confidence. The constant input uncertainty assumption used for training the model indicated that we are equally confident (we use a low value of $0.01$) about every feature. We propose to quantify how much the model accumulates additional uncertainties to each of the input features, as the prediction uncertainty grows. We use the trained model $f(\boldsymbol{\theta})$ to adjust the input uncertainties by artificially increasing the prediction variance, without affecting the mean estimates. To achieve this, we use the KL-divergence loss to update $\sigma_i^{(j)}$'s, while keeping the network parameters frozen. It is important to note that, the updated uncertainties do not provide a description of the real-world. Instead, this is the model's hypothesis of how the prediction uncertainty can be potentially decomposed into the uncertainties at each of the input features. 

For a given data sample $\mathcal{N}(\mathbf{x},\boldsymbol{\sigma})$ with prediction $\mathcal{N}(\hat{y},\beta)$, let us denote the set of estimated input uncertainties for each feature $j$, when the prediction variance increases, as $\{\hat{\sigma}_t^{(j)}\}$. Here, $t$ denotes the desired factor of increase in $\beta$. In our experiments, we set $t$ at $\{1.1, 1.25, 1.5, 1.75, 2.0, 2.5\}$. We propose a novel \textit{uncertainty gap} score indicating which input features, according to the model, maximally contribute to the prediction variance. The score is given by:

\begin{equation}
\text{gap}(x_i^{(j)}) = \text{AUC}\left([\beta_t],[\hat{\sigma}_t^{(j)}]\right),
\label{eqn:gap}
\end{equation}where $\text{AUC}$ indicates the area under the curve metric, measured from the plot for prediction variances vs. estimated input uncertainty, for different values of $t$. 

%



\begin{figure*}[!htb]
	\centering
	\subfigure[Feature ranking based on sensitivities]{\includegraphics[width=0.33\linewidth]{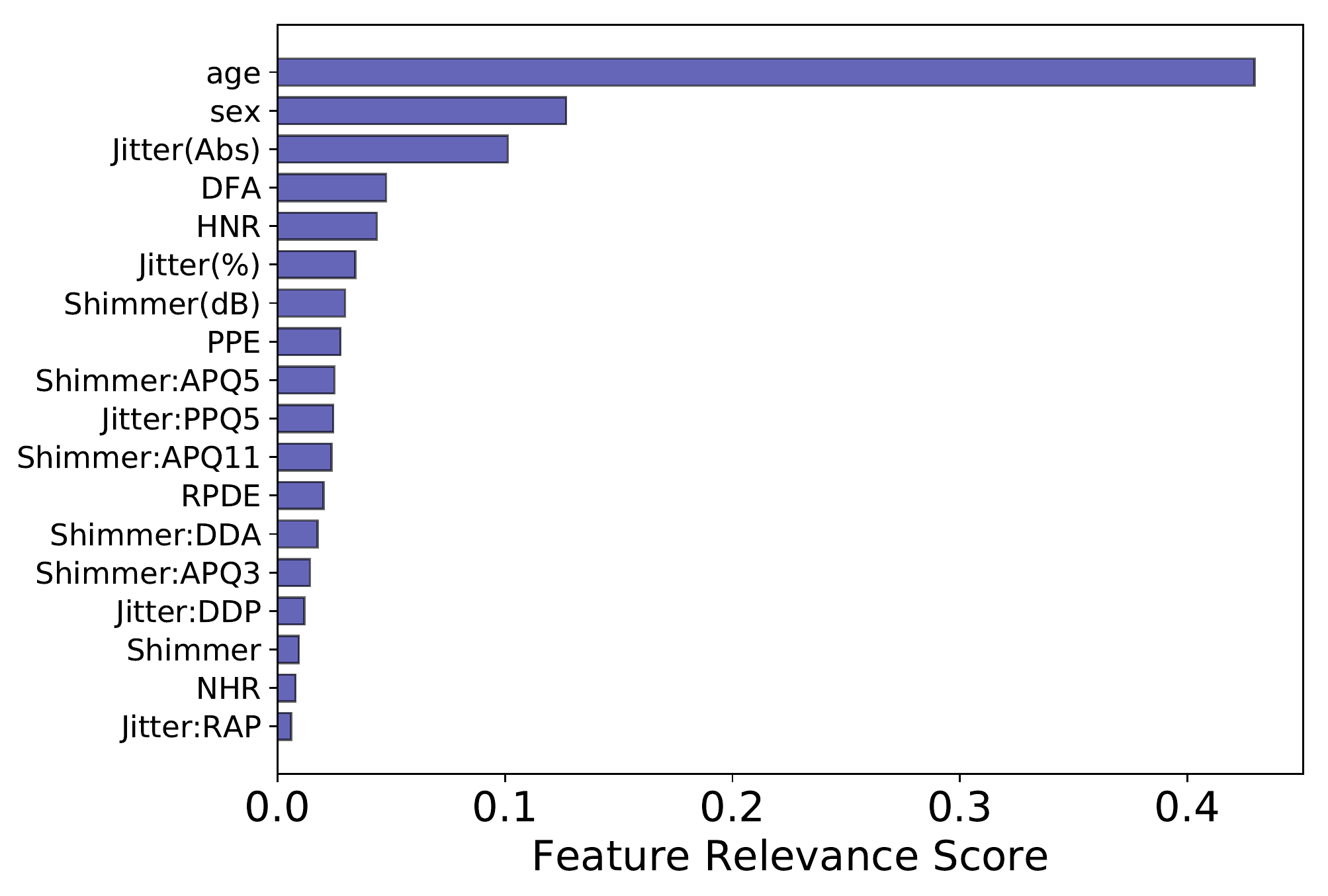}	\label{fig:park0}}
	\subfigure[Impact of masking the less sensitive features]{\includegraphics[width=0.3\linewidth]{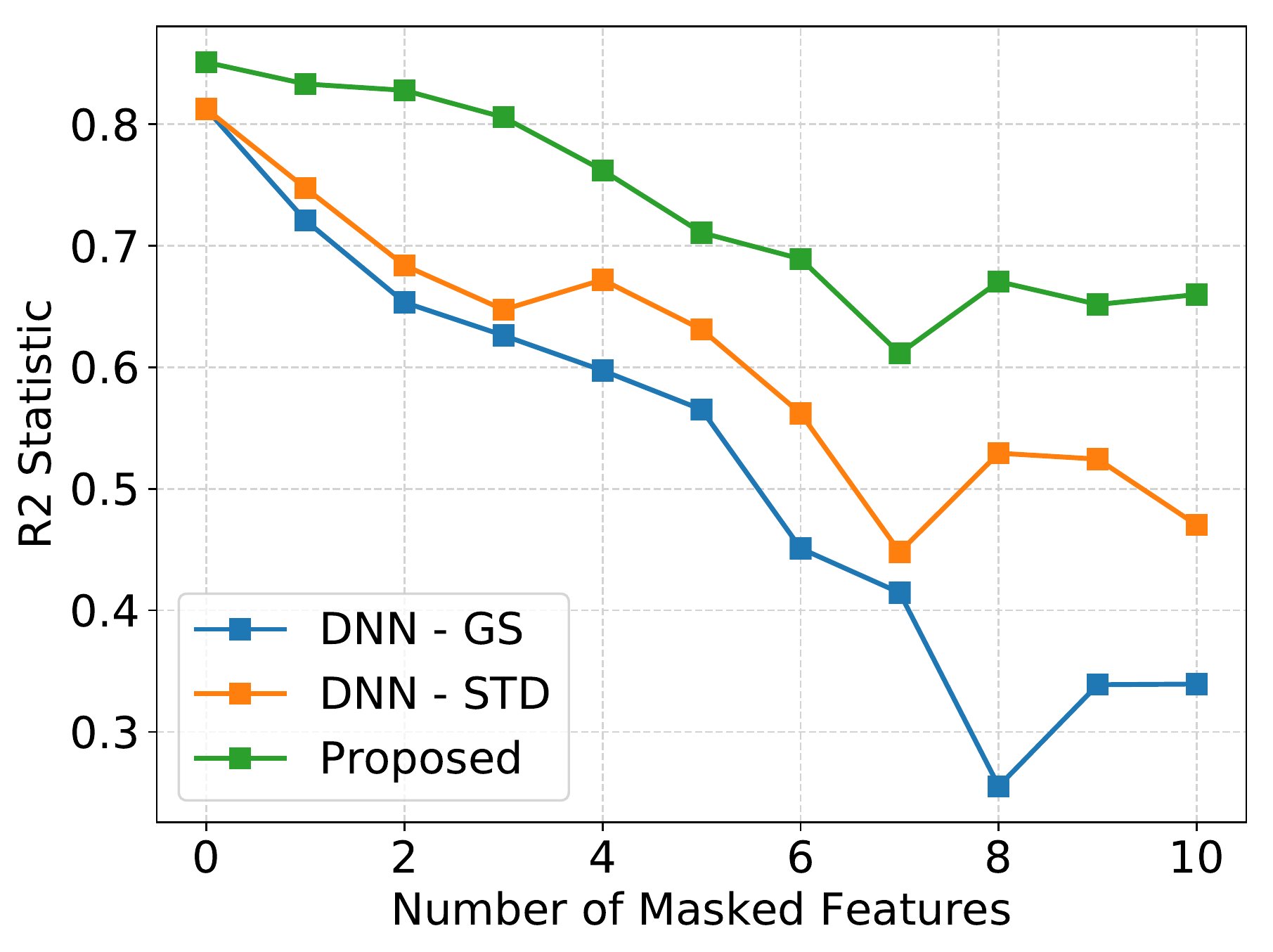}\label{fig:park1}}
	\subfigure[Estimated prediction uncertainties]{\includegraphics[width=0.3\linewidth]{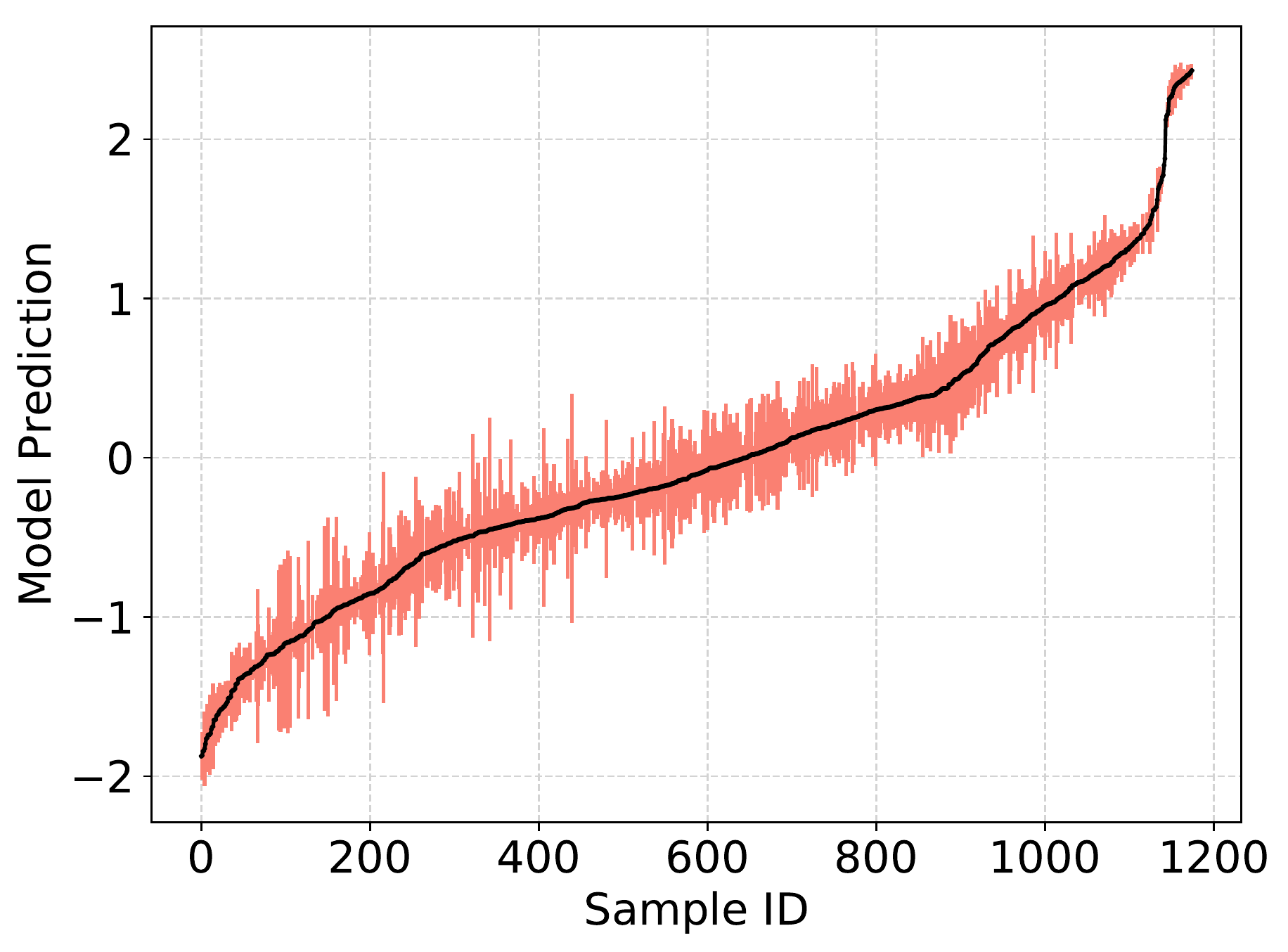}\label{fig:park2}}	
	\label{fig:park_sens}
	
	\subfigure[]{\includegraphics[width=0.33\linewidth]{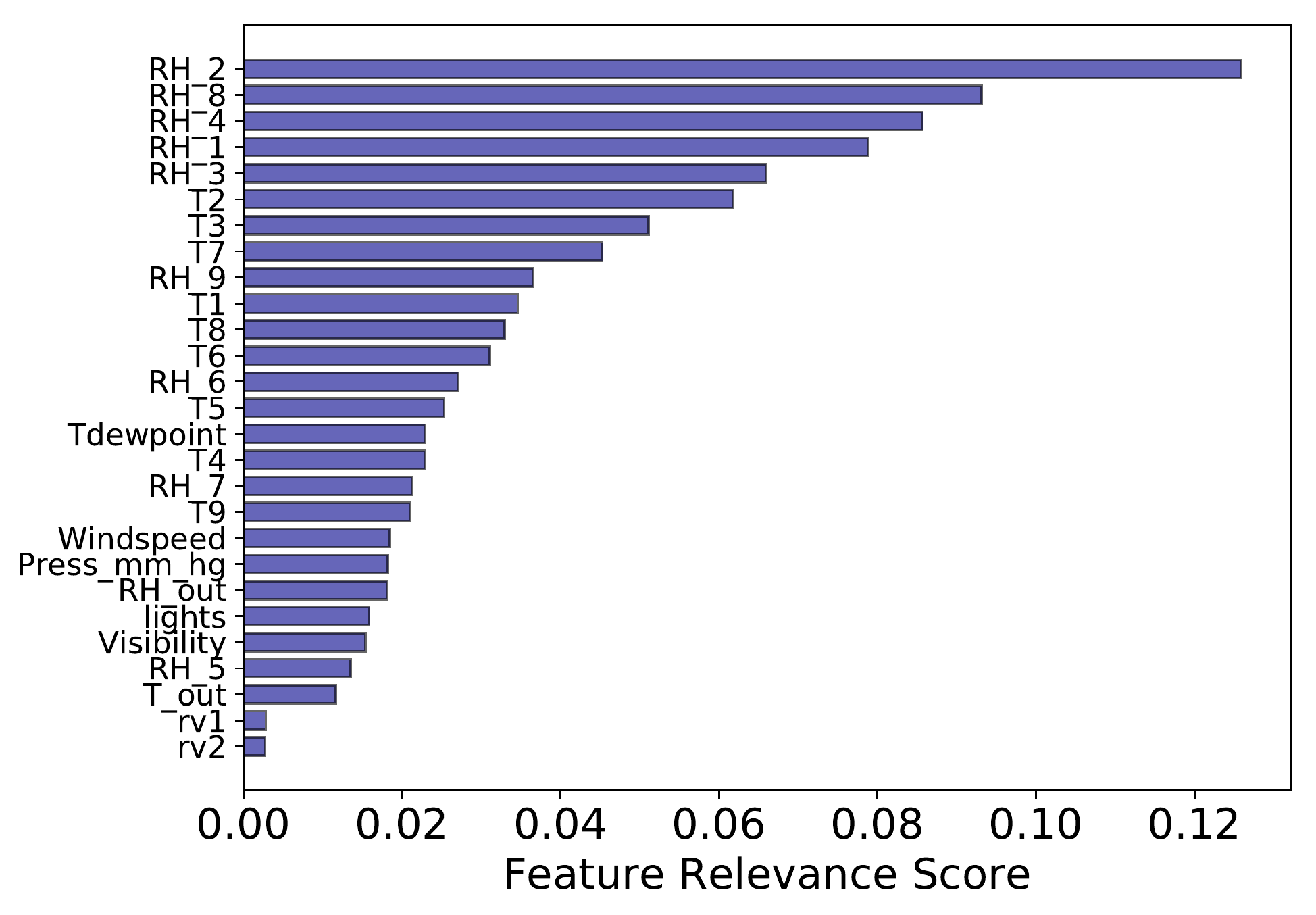}\label{fig:energy0}}
	\subfigure[]{\includegraphics[width=0.3\linewidth]{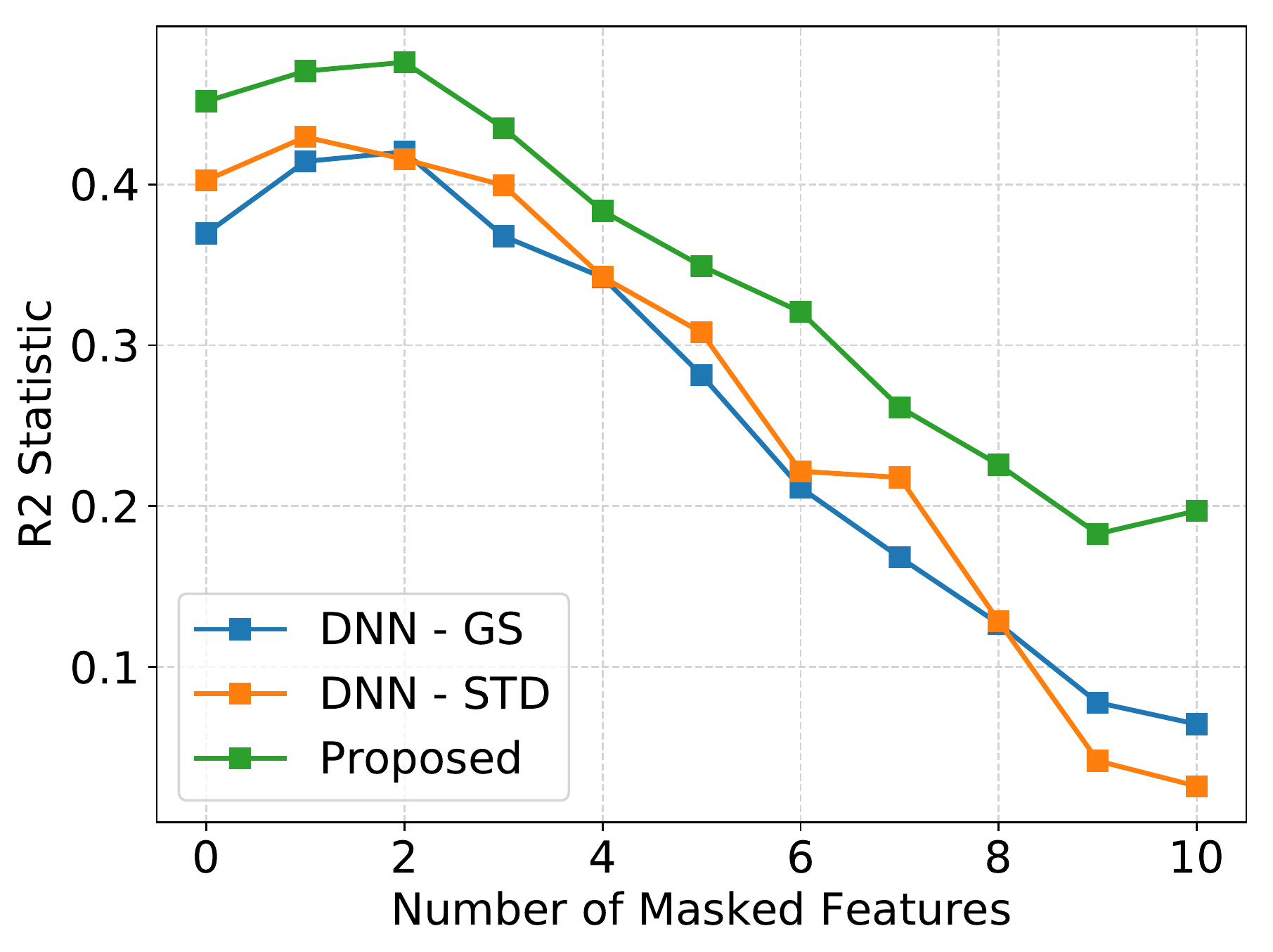}\label{fig:energy1}}
	\subfigure[]{\includegraphics[width=0.3\linewidth]{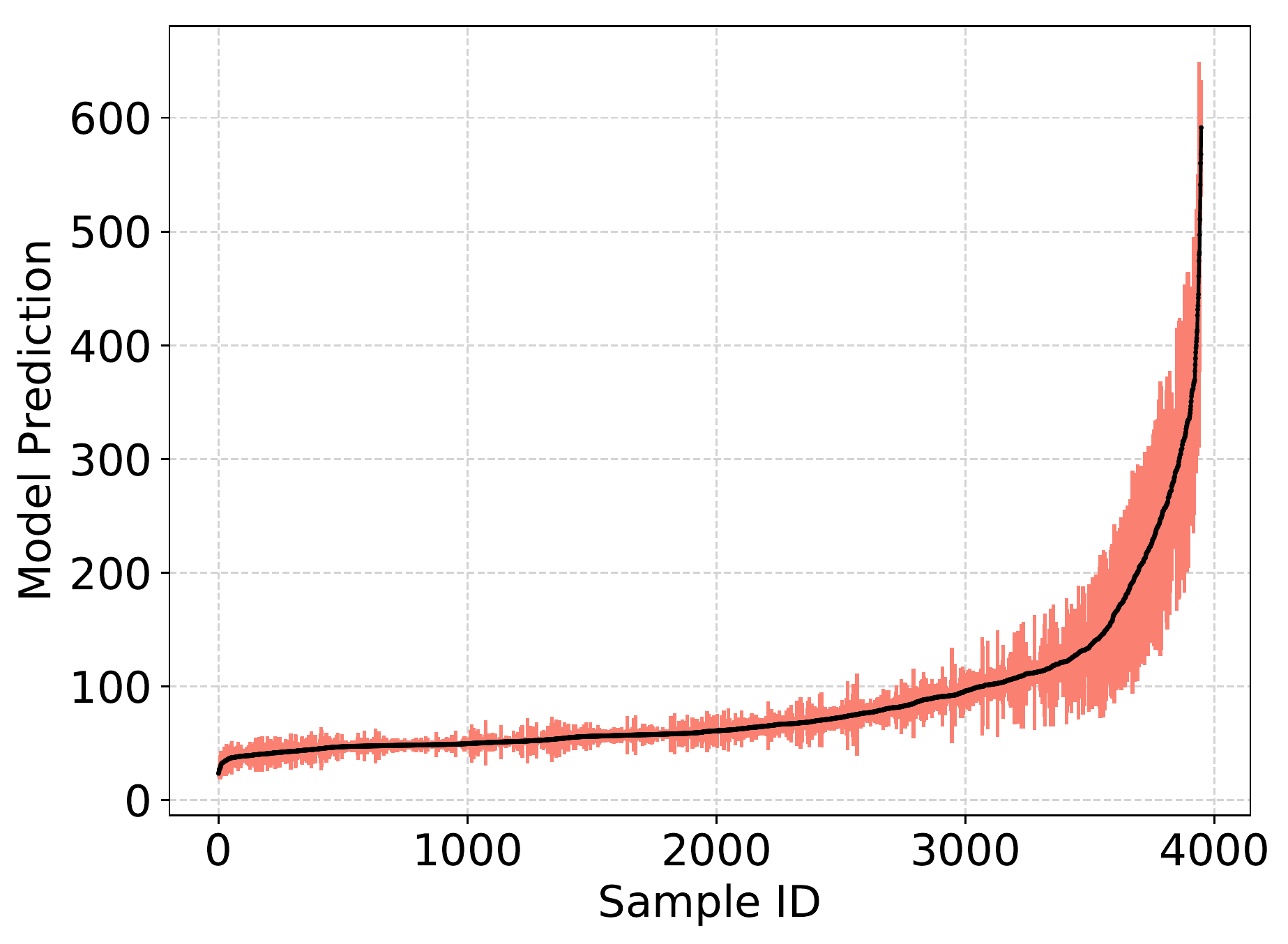}\label{fig:energy2}}	
	\vspace{-0.1in}
	\caption{Performance of the predictive model obtained using the proposed approach on the \textit{Parkisons Telemonitoring} (top row) and \emph{Appliance Energy Usage} (bottom row) datasets.}
\end{figure*}

\begin{figure}[!htb]
	\centering
	\subfigure[Low UPDRS case]{\includegraphics[width=0.49\linewidth]{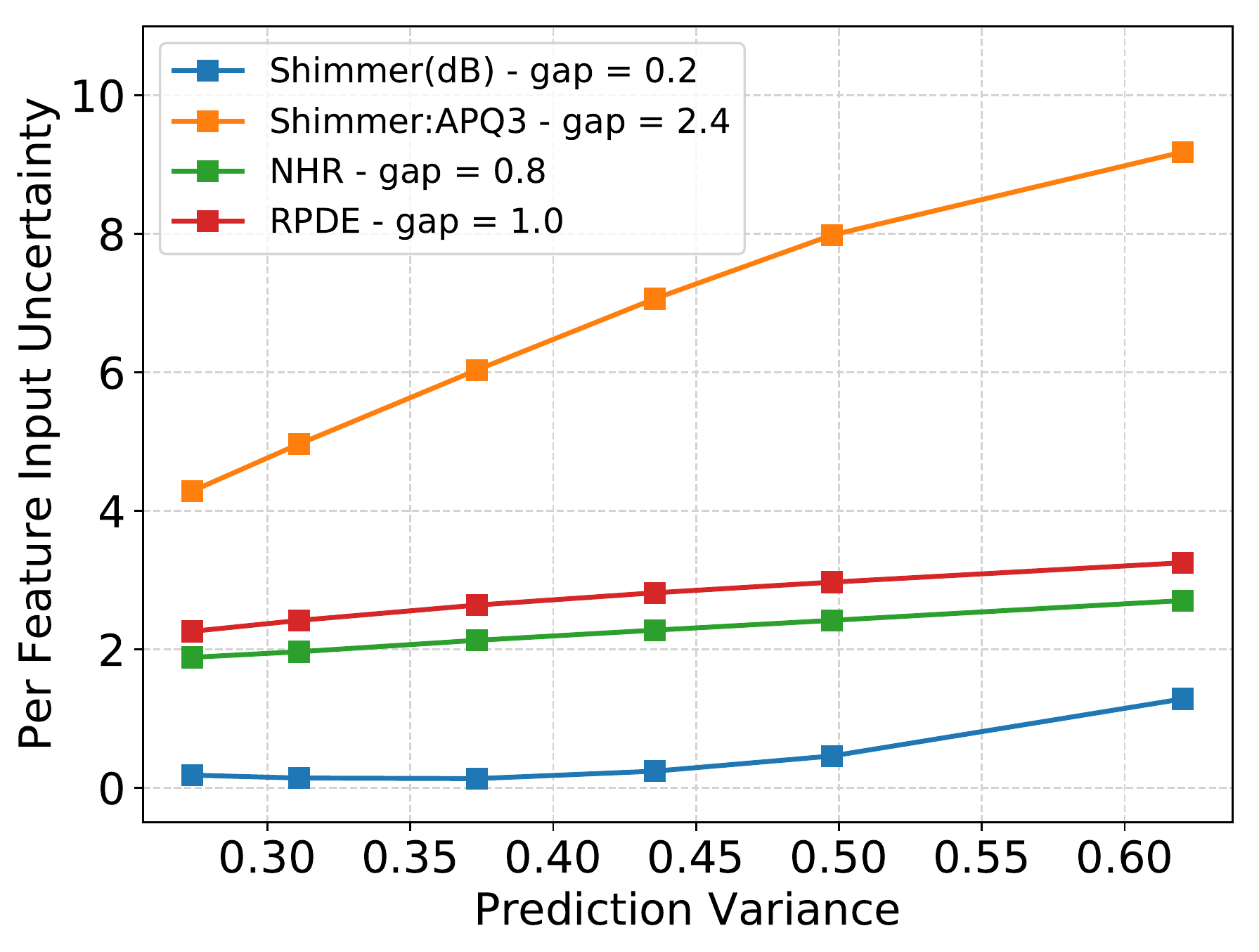}}
	\subfigure[High UPDRS case]{\includegraphics[width=0.49\linewidth]{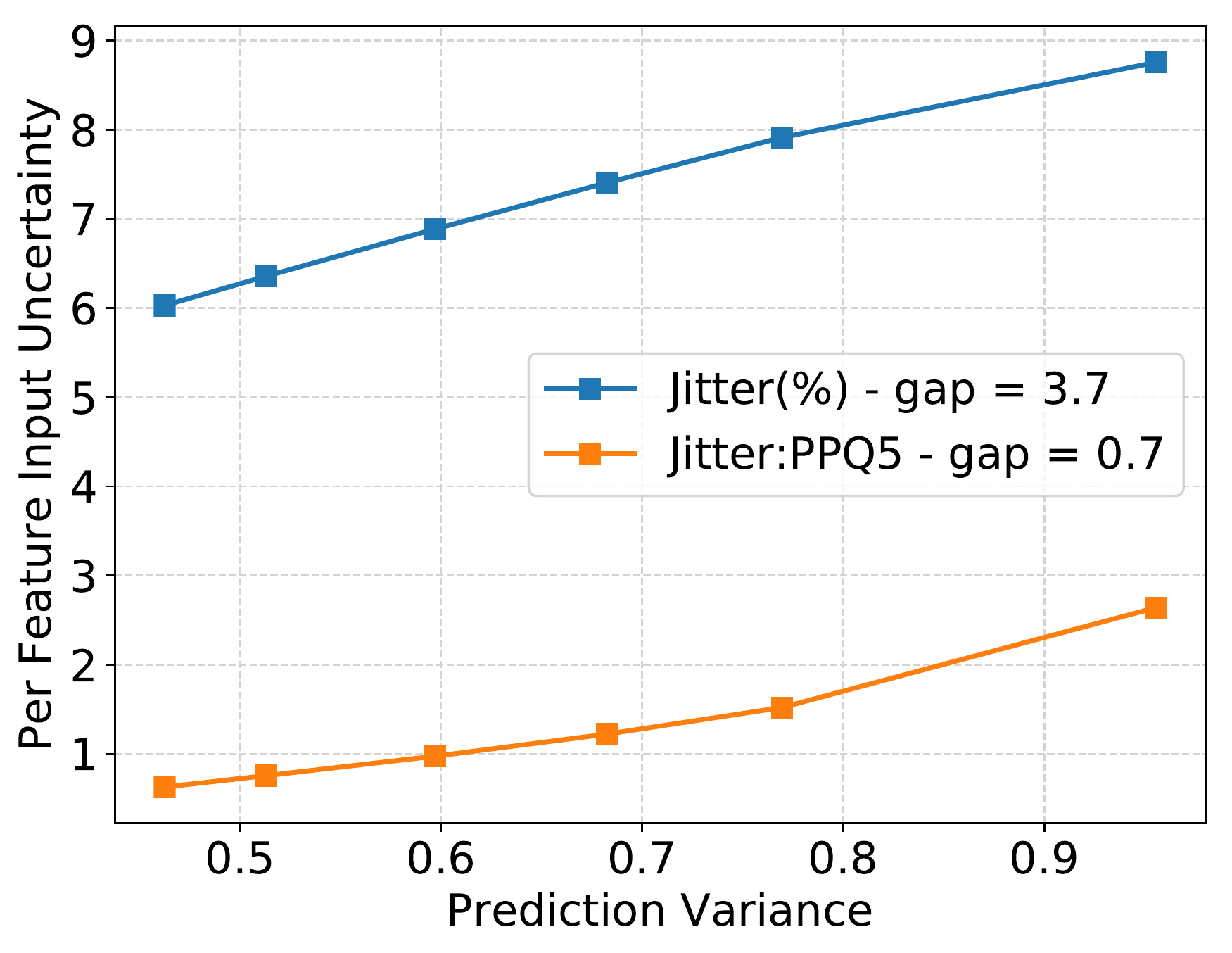}}
	\vspace{-0.1in}
	\caption{\textit{Uncertainty gap} scores for different input features obtained with two examples from the \textit{Parkisons} dataset.}
	\label{fig:park_gap}
\end{figure}

%% file: exps.tex

In this section, we perform experiments with two standard regression datasets, by employing the proposed approach for gaining insights into the predictive model.

\noindent \textbf{Datasets}: (i) \textit{Parkinsons Telemonitoring Dataset}~\cite{tsanas2010accurate} - This dataset is comprised of a range of biomedical voice measurements from subjects with early-stage Parkinson's disease recruited for testing a telemonitoring device for remote symptom progression monitoring. In particular, there are a total of $18$ measurements (e.g. measures of variation in fundamental frequency, measures of variation in amplitude) corresponding to each of $5,875$ recordings. The goal is to predict the UPDRS (Unified Parkinson’s Disease Rating Scale) score~\cite{ivey2012unified}, that indicates the disease severity. (ii) \textit{Appliance Energy Usage Dataset:}~\cite{candanedo2017data} - This dataset contains measurements pertinent to house temperature and humidity conditions and the goal is to estimate the amount of energy usage by the appliances (in Wh units). There are $29$ input attributes, out of which two of them are random variables, corresponding to $19,735$ samples. 	

\noindent \textbf{Experiment Setup}: In both datasets, we trained LPN models with $4$ dense layers of sizes $256-128-16-1$ along with leaky ReLU activation and dropout with $p=0.3$. The models were trained using the Adam optimizer with learning rate $0.0005$. We trained the models using $80\%$ of the data and validated using the remaining $20\%$, and the reported results were obtained using cross validation.

\noindent \textbf{Results}: Figure \ref{fig:park0} ranks the $18$ input features from the Parkinsons dataset, based on their relevance scores, while Figure \ref{fig:park1} shows the impact of perturbing less relevant features (at test time) on the prediction performance. More specifically, we incrementally mask one feature at time (low to high in relevance) by replacing that feature with a constant value (set to median value of the train data) and measure the $R^2$ (R-squared) statistic on the validation dataset. For comparison we show similar results obtained using standard neural networks, with the same architecture, wherein the feature ranks were obtained using gradient based sensitivities (DNN - GS)~\cite{montavon2017methods} and simple Taylor decomposition (DNN-STD)~\cite{bach2015pixel}. The first observation is that, the proposed approach produces improved validation performance compared to models that did not take uncertainties into account. A more surprising observation is the amount of  performance degradation, when less relevant features are masked, is significantly lower with our model. 

Figure \ref{fig:park2} illustrates the predictions obtained from our model, along with their uncertainties. Following our approach in Section 3.3, we can utilize the \textit{uncertainty gap} score to understand the prediction uncertainties in terms of the input features. Figure \ref{fig:park_gap} shows the gap scores for two different samples (one with low and other with high UPDRS scores). With subjects that have low UPDRS, the model finds the \textit{Shimmer APQ3} to be the major source for uncertainties in prediction. On the other hand, for a patient with higher degree of severity, \textit{Jitter (\%)} is the major source of uncertainty. A domain expert can compare these estimates with their modeling of the physical world to evaluate the fidelity of the model. For example, if the model had picked the \textit{age} or the \textit{sex} variables as the prominent sources of uncertainty, this model would be suspicious, as there is no reason to believe there can be high uncertainties about those variables.

We obtain similar observations with the \textit{Appliance Energy Usage} dataset, as shown in Figures \ref{fig:energy0},\ref{fig:energy1},\ref{fig:energy2}. As expected, the model rejects both the random attributes by assigning low relevance scores, while the humidity parameters are found to be more relevant compared to external factors such as \textit{windspeed} or \textit{visibility}. Similar to the previous example, Figure \ref{fig:energy1} demonstrates improvements in validation performance as we perturb the less relevant features. Further, Figure \ref{fig:energy2}, we observe that the prediction variance for samples with high energy usage is much larger than those for lower values. Upon close investigation of the gap scores, we made a surprising observation that the random variables were the culprits, and retraining the model without them ended up shrinking the prediction variances significantly.